\documentclass[sigconf]{acmart}

\AtBeginDocument{%
  }

\setcopyright{acmlicensed}
\copyrightyear{2026}
\acmYear{2026}
\acmDOI{XXXXXXX.XXXXXXX}
\acmConference[MM '26]{the 34th ACM International Conference on Multimedia}{Nov. 10--14, 2026}{Rio de Janeiro, Brazil}
\acmISBN{978-1-4503-XXXX-X/2026/11}

\usepackage{multirow}
\usepackage{graphicx}
\usepackage{booktabs}
\usepackage{enumitem}

\begin{document}

\title{AerialClaw: An Open-Source Framework for LLM-Driven Autonomous Aerial Agents}


\author{Ke Li}
\affiliation{%
  \institution{Xidian University}
  \city{Xi'an}
  \country{China}}
\email{like0413@stu.xidian.edu.cn}

\author{Jianfei Yang}
\affiliation{%
  \institution{Xidian University}
  \city{Xi'an}
  \country{China}}
\email{yjf123@stu.xidian.edu.cn}

\author{Luyao Zhang}
\affiliation{%
  \institution{Xidian University}
  \city{Xi'an}
  \country{China}}
\email{lyzhang_5@stu.xidian.edu.cn}

\author{Guo Yu}
\affiliation{%
  \institution{Xidian University}
  \city{Xi'an}
  \country{China}}
\email{guoyu1999@stu.xidian.edu.cn}

\author{Chengwei Yan}
\affiliation{%
  \institution{Xidian University}
  \city{Xi'an}
  \country{China}}
\email{ycw@stu.xidian.edu.cn}

\author{Yuan Ding}
\affiliation{%
  \institution{Xidian University}
  \city{Xi'an}
  \country{China}}
\email{25031110055@stu.xidian.edu.cn}

\author{Di Wang}
\authornote{Corresponding author.}
\affiliation{%
  \institution{Xidian University}
  \city{Xi'an}
  \country{China}}
\email{wangdi@xidian.edu.cn}

\author{Nan Luo}
\affiliation{%
  \institution{Xidian University}
  \city{Xi'an}
  \country{China}}
\email{nluo@xidian.edu.cn}

\author{Gang Liu}
\affiliation{%
  \institution{Xidian University}
  \city{Xi'an}
  \country{China}}
\email{gliu@xidian.edu.cn}

\author{Xiao Gao}
\affiliation{%
  \institution{Xi'an University of Architecture and Technology}
  \city{Xi'an}
  \country{China}}
\email{gaoxiao@xauat.edu.cn}

\author{Quan Wang}
\affiliation{%
  \institution{Xidian University}
  \city{Xi'an}
  \country{China}}
\email{qwang@xidian.edu.cn}

\begin{abstract}
Unmanned aerial vehicles (UAVs) are increasingly used in inspection, search and rescue, environmental monitoring, and emergency response. However, most UAV applications still rely on pre-defined command sequences or task-specific pipelines, where developers manually connect perception, planning, flight control, simulation, logging, and safety modules. This limits the flexibility, reproducibility, and extensibility of autonomous aerial systems. This paper presents AerialClaw, an open-source software framework that enables UAVs to operate as decision-making aerial agents rather than merely command-following platforms. Given a natural-language mission, AerialClaw allows an LLM-based agent to understand the task, maintain context, invoke executable aerial skills, observe perception and runtime feedback, and iteratively update its decisions in a closed loop. The framework adopts a modular \emph{brain–skill–runtime} architecture, combining hard skills for atomic UAV operations, Markdown-based soft skills for reusable task strategies, document-driven agent state and capability boundaries, memory-driven reflection, safety-oriented runtime validation, and platform-agnostic execution adapters. AerialClaw supports lightweight mock execution, PX4 SITL with Gazebo, and AirSim-based simulation, together with a web console, pluggable model backends, example missions, simulation assets, and staged deployment scripts. By combining standardized aerial skills, document-driven agent state, memory, and closed-loop LLM decision-making, AerialClaw provides a reproducible and extensible open-source framework for building UAV systems that can interpret missions, make decisions, execute skills, and adapt their behavior from feedback.
\end{abstract}

\ccsdesc[500]{Computing methodologies~Artificial intelligence}

\keywords{Open-source software, autonomous aerial agents, UAV systems, large language models, embodied agents}

\maketitle

\noindent\textbf{Relevant ACM MM areas:} Embodied and Immersive Multimedia; Multimedia Applications.

\noindent\textbf{Open-source availability:}
AerialClaw is publicly available at \textcolor{blue}{\url{https://github.com/XDEI-Group/AerialClaw}}. The repository provides the source code, open-source license, documentation, build instructions, example missions, simulation assets, and configuration files. The supplementary ZIP archive submitted to the ACM MM Open Source Software track packages the same release materials for offline inspection.

\begin{figure}[!t]
    \centering
    \includegraphics[width=\linewidth]{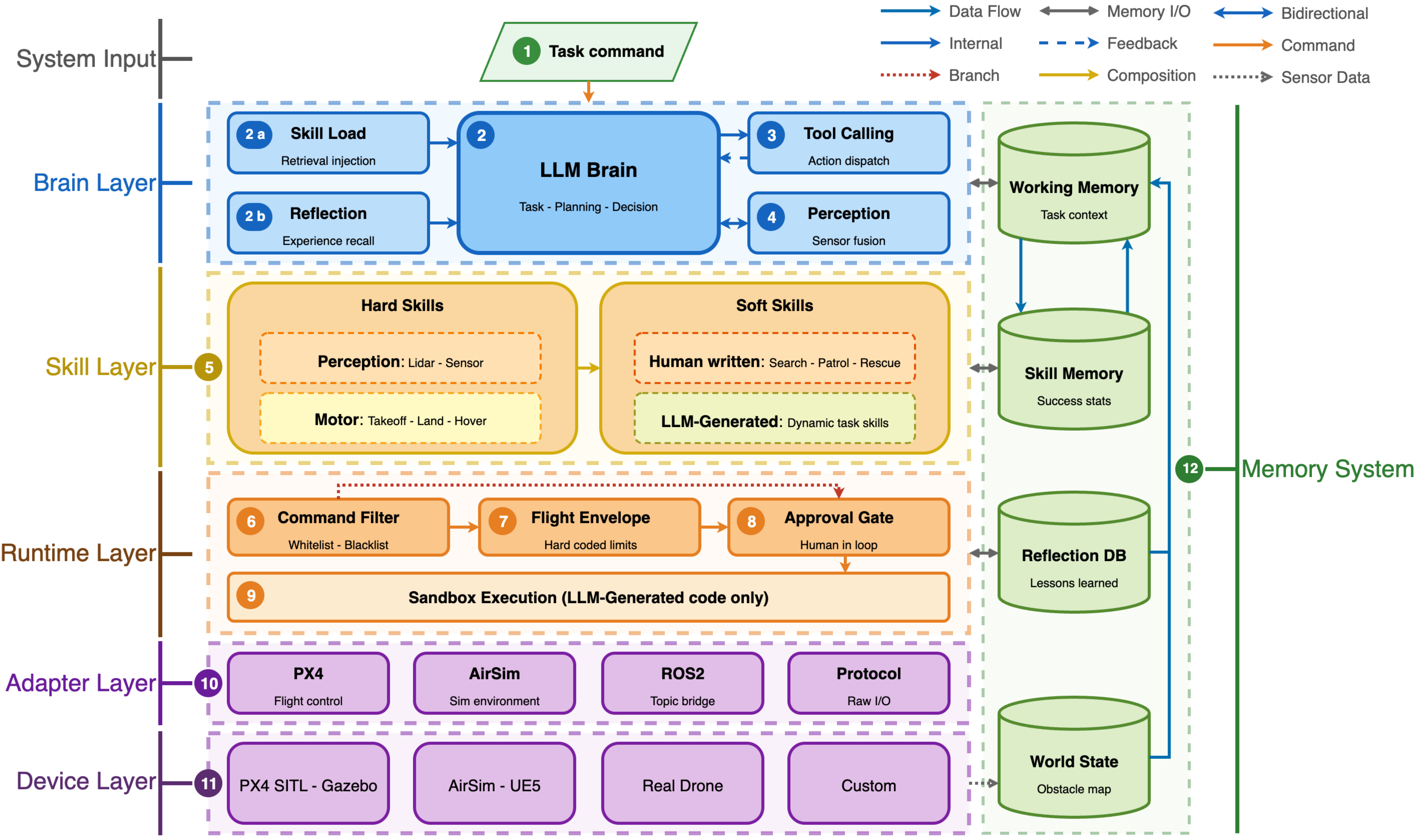}
    \caption{The overall architecture of AerialClaw. The framework connects an LLM-based cognitive agent, hybrid skill system, memory modules, perception pipeline, safety-aware runtime, and platform-specific adapters for mock execution, PX4/Gazebo, AirSim, and custom UAV backends.}
    \label{fig:architecture}
\end{figure}

\section{Introduction}
\label{sec:intro}
Unmanned aerial vehicles (UAVs) have become important platforms for inspection, search and rescue, environmental monitoring, emergency response, and infrastructure maintenance. However, most UAV applications are still developed around pre-defined command sequences or task-specific pipelines. Developers usually specify waypoints, write control logic, connect perception modules, and manually handle runtime states for each mission. As a result, the UAV largely acts as a command-following platform rather than an autonomous agent that can understand a mission, reason about context, and decide what to do next.

Recent advances in large language models (LLMs) and embodied agents suggest a new way to build aerial autonomy\cite{zhao2026survey,huo2026abot,openclaw2026,ravichandran2025spine,cladera2025airground,liu2025hierarchical}. Instead of pre-scripting every mission procedure, a UAV can be equipped with an agentic control loop: it receives a natural-language mission, decomposes the task, selects executable skills, observes perception and runtime feedback, and updates its decisions during execution. To make this paradigm practical, developers need software infrastructure that connects high-level agent reasoning with aerial skills, simulator or hardware backends, memory, runtime validation, and user-facing control tools. Existing UAV ecosystems support flight control, simulation, communication, and sensor integration, but they do not directly provide a reusable runtime for LLM-driven aerial agents. Developers therefore still need to rebuild integration logic for each new mission, simulator, or UAV platform, which limits reuse and reproducibility.

We present AerialClaw, an open-source software framework for constructing LLM-driven autonomous aerial agents. AerialClaw moves UAV applications beyond pre-defined command execution by enabling UAV systems to operate as decision-making aerial agents. Given a natural-language mission, an LLM-based agent maintains context, invokes executable aerial skills, observes feedback, and iteratively decides the next action. The framework adopts a modular \emph{brain-skill-runtime} architecture: the brain handles task understanding, planning, decision generation, and reflection; the skill system combines hard skills for atomic UAV operations with Markdown-based soft skills for reusable task strategies; and the runtime validates and executes decisions through platform-agnostic adapters for mock execution, PX4\cite{meier2015px4}/Gazebo\cite{koenig2004design}, AirSim\cite{shah2017airsim}, or custom UAV backends.

A key design principle of AerialClaw is to make aerial-agent behavior inspectable and extensible. Robot identity, hardware capabilities, available skills, task strategies, and accumulated experience are represented through human-readable documents. This document-centered interface allows developers to adjust agent behavior, add new task strategies, and extend platform capabilities without retraining models or rewriting the core runtime. AerialClaw also provides a web console for submitting missions and monitoring decisions, telemetry, logs, and runtime feedback. The public release includes source code, documentation, example missions, simulation assets, configuration files, deployment scripts, and an open-source license.
The main contributions of this work are as follows:
\begin{itemize}[leftmargin=*]
    \item A decision-making framework for autonomous aerial agents. AerialClaw enables UAV systems to move beyond pre-scripted command execution by integrating natural-language mission understanding, closed-loop decision-making, executable skills, perception feedback, and runtime validation.
    \item A hybrid skill and document-centered agent interface. AerialClaw combines hard skills for atomic UAV operations with Markdown-based soft skills for reusable task strategies, while representing agent identity, capability boundaries, and memory through human-readable documents.
    \item A platform-agnostic open-source execution stack. AerialClaw decouples high-level agent reasoning from simulator- and hardware-specific control through a unified runtime and adapter interface, supporting mock execution, PX4/Gazebo, AirSim, and custom UAV backends.
\end{itemize}

\section{Software Overview}
\label{sec:overview}
Figure~\ref{fig:architecture} shows the architecture of AerialClaw. The framework is organized as a modular ``brain--skill--runtime'' stack that turns a natural-language mission into closed-loop UAV behavior. Rather than executing a pre-defined command sequence, AerialClaw lets an LLM-based agent repeatedly understand the current task context, select an executable skill, observe feedback, and decide the next action. The system consists of four main components: an LLM Agent, a Skill System, an Execution Interface, and a Memory System.

\subsection{LLM Agent}
The LLM Agent is responsible for task understanding, step-wise planning, and decision generation. At each iteration, it receives a structured context that includes the user mission, available skills, perception summary, world state, execution history, and retrieved experiences. It then outputs an inspectable JSON decision specifying the next skill call and its arguments. The runtime validates this decision before execution, and the resulting observation is fed back to the agent for the next step. This closed-loop design allows the UAV to adapt its behavior during execution instead of following a fixed plan generated in advance.

\subsection{Document-centered Interface}
AerialClaw represents robot identity, capability boundaries, task strategies, and accumulated experience through human-readable documents, as shown in Fig.~\ref{fig:doc_interface}. \texttt{SOUL.md} defines the agent's identity and operating principles, while \texttt{BODY.md} is generated from the active adapter and sensor configuration at startup. This keeps the agent's prompt aligned with the current simulator or hardware setup, including supported skills, sensor availability, flight limits, and battery-related constraints. Strategy documents under \texttt{SKILLS/*.md} describe reusable soft skills, and \texttt{MEMORY.md} records lessons and feedback from previous missions. This interface makes agent behavior inspectable and editable without retraining models or modifying the core runtime.

\begin{figure}[t]
\centering
\small
\fbox{%
\begin{minipage}{0.95\linewidth}
\textbf{Document-centered agent interface}\\[2pt]
\texttt{SOUL.md}: identity, communication style, operating principles\\
\texttt{BODY.md}: active sensors, flight limits, available skills\\
\texttt{SKILLS/*.md}: reusable task strategies and constraints\\
\texttt{MEMORY.md}: lessons, environment insights, skill feedback
\end{minipage}}
\caption{AerialClaw exposes agent behavior through human-readable documents, allowing users to inspect and modify identity, capability, strategy, and experience without changing the core runtime.}
\label{fig:doc_interface}
\end{figure}

\subsection{Skill System}
The Skill System grounds high-level reasoning in executable UAV behavior. AerialClaw distinguishes between \emph{hard skills} and \emph{soft skills}. Hard skills are callable functions for atomic operations such as takeoff, waypoint flight, observation, object detection, status query, map marking, and return-to-launch. Soft skills are Markdown-based strategy documents that describe reusable task procedures, such as target search, area patrol, rescue workflow, and inspection routine. During execution, the LLM Agent retrieves relevant soft-skill documents and composes hard skills according to the current mission context. Developers can therefore extend low-level platform capabilities by adding hard skills, or modify high-level task behavior by editing soft skills.

\subsection{Execution Interface}
The Execution Interface decouples agent decisions from simulator- or hardware-specific control. The Runtime Layer validates each skill call using safety-oriented checks such as skill allowlists, parameter constraints, geofencing, human approval for critical commands, and sandboxing for dynamically generated code. The Adapter Layer then translates validated abstract skills into backend-specific commands. Current adapters support lightweight mock execution, PX4 SITL through MAVSDK/MAVLink\cite{koubai2019mavlink}, and AirSim through its RPC interface, while the same interface can be extended to real UAVs or custom platforms. This design allows the same agent logic to run across different execution backends.

\subsection{Memory System and Model Backends}
The Memory System provides context for decision-making and experience reuse. It includes working memory for current task state and execution history, skill memory for recent skill status and usage statistics, a reflection database for post-task lessons and semantic retrieval, and world state for robot status, known obstacles, marked targets, and environment information. After a mission, the Reflection Engine summarizes the execution trace into reusable lessons, which can later be retrieved for similar tasks or converted into new soft-skill documents. AerialClaw also uses an OpenAI-compatible client abstraction for LLM and VLM calls, allowing users to switch between cloud APIs, local models, and custom services through configuration.

\section{Main Features and Applications}
\label{sec:features}

AerialClaw is designed for researchers and developers who need a reusable platform for building autonomous aerial agents. Its main features include natural-language mission execution, closed-loop perception--reasoning--action, hybrid hard/soft skills, memory-driven reflection, cross-backend deployment, and a web-based control console. Instead of requiring users to predefine a complete mission procedure, AerialClaw allows the agent to interpret a high-level instruction, select executable skills, observe feedback, and decide the next action during execution.

Through the web console, users can submit missions, monitor agent decisions, inspect execution logs, view telemetry and map state, and observe sensor outputs. This makes the decision process visible during operation and helps developers debug the interaction between reasoning, skill execution, perception, and runtime validation.

AerialClaw supports application scenarios such as area reconnaissance, facility inspection, target search, environmental monitoring, emergency response, and education-oriented UAV agent prototyping. For multimedia and embodied-AI research, it provides a software testbed for connecting language instructions, aerial visual observations, spatial state, and executable UAV actions. For system developers, it provides extensible interfaces for adding new sensors, simulators, UAV adapters, model backends, and task skills.

\section{Implementation and Reproducibility}
\label{sec:implementation}
AerialClaw is implemented as an open-source Python framework with a web-based control interface. The backend uses Flask and Socket.IO for real-time bidirectional communication between the runtime and the user interface. The frontend is built with React and visualizes agent decisions, execution logs, telemetry, map state, and sensor outputs. Model calls are handled through an OpenAI-compatible interface, allowing planning, visual analysis, and reflection modules to use cloud APIs, local models, or custom services through configuration.

For execution backends, AerialClaw provides adapters for lightweight mock execution, PX4 SITL with Gazebo, and AirSim. The mock backend supports quick inspection of the agent loop without UAV hardware or full simulator dependencies. The PX4/Gazebo backend supports physics-based simulation through PX4 Autopilot, Gazebo, MAVSDK, and sensor-equipped UAV models. The AirSim backend is provided for camera-rich aerial scenes and VLM-based perception experiments. The repository includes the Python backend, React frontend, hard-skill implementations, Markdown-based soft-skill documents, simulator adapters, example configurations, technical documentation, and Gazebo assets.

\subsection{Deployment and Quick Start}
AerialClaw provides a progressive deployment path from software-only execution to embodied simulation while preserving the same agent architecture. The lightweight entry point is a \textbf{containerized mock runtime}, which launches the Web console, backend service, in-memory UAV adapter, skill interface, and agent loop without UAV hardware, PX4, Gazebo, AirSim, Unreal Engine, or GPU dependencies. Users can then enable the \textbf{LLM-driven control loop} through an OpenAI-compatible endpoint, and further connect the same codebase to \textbf{PX4 SITL + Gazebo} for physics-based simulation with PX4 Autopilot, Gazebo, MAVSDK, and sensor-equipped UAV models. \textbf{AirSim + Unreal Engine} is maintained as an optional photorealistic extension for camera-rich aerial scenes and VLM-based perception experiments.

\subsubsection{Containerized quick start.}
For lightweight and reproducible execution, AerialClaw provides a Docker entry point that starts the core software stack with a mock UAV adapter:
{\color{blue!70!black}
\begin{verbatim}
git clone https://github.com/XDEI-Group/AerialClaw.git
cd AerialClaw
docker compose up
\end{verbatim}
}
The Web console is available at \url{http://localhost:5001}, and \texttt{/api/status} provides a basic health check. This mode exposes natural-language task submission, skill invocation, execution tracing, memory state, and human override while avoiding simulator-specific installation requirements.

\subsubsection{PX4/Gazebo quick start.}
The containerized mock runtime and the PX4/Gazebo simulator path are alternative launch modes sharing the default Web port \texttt{5001}. Before starting the simulator stack, the mock container should be stopped:
{\color{blue!70!black}
\begin{verbatim}
docker compose down
./scripts/sim_quickstart.sh --setup
./scripts/sim_quickstart.sh
\end{verbatim}
}
The quickstart script orchestrates PX4 SITL, Gazebo, the AerialClaw backend, Web UI, Gazebo GUI, MAVSDK control link, and camera/LiDAR sensor bridge. It also performs built-in health checks for the Web service, PX4 adapter, sensor bridge, and camera endpoint, preventing silent fallback from simulator control to mock control. Through the adapter interface, the high-level mission logic remains unchanged across mock and physics-based deployments.

\subsection{Installation support and availability}
AerialClaw follows the Open Source Software track recommendation of easing build and testing through containers and package managers. The release provides Docker-based quick start commands, Python backend dependency installation, npm-based Web UI build instructions, and a guided PX4/Gazebo setup script. The optional AirSim/Unreal path is documented separately and is not required for the main open-source release. The public repository contains the source code, MIT license, documentation, example configurations, and build/install instructions. The supplementary archive packages the same release materials for offline inspection and reproducibility checking.

\begin{figure}[!t]
    \centering
    \includegraphics[width=\linewidth]{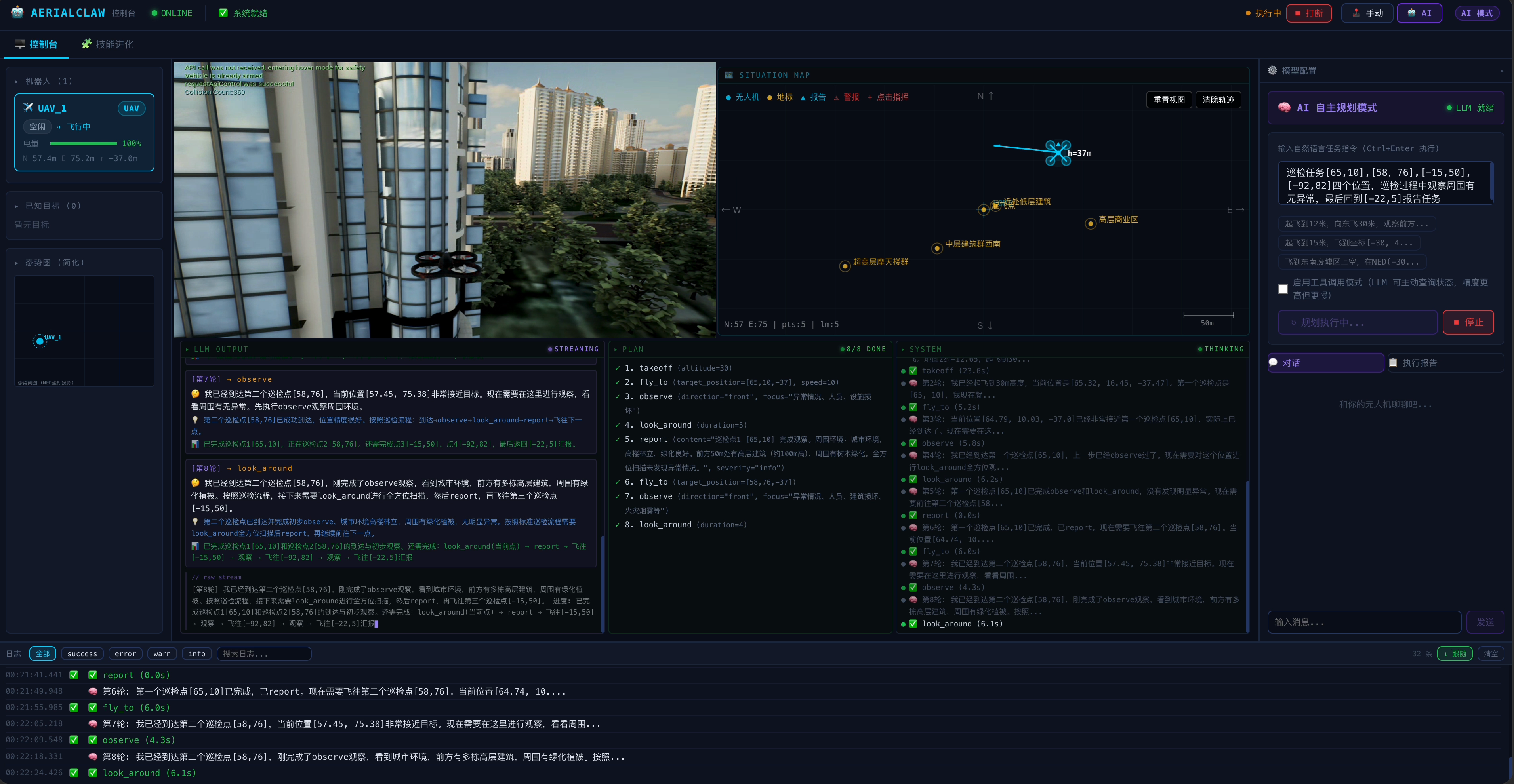}
    \caption{AerialClaw web console for monitoring an autonomous aerial inspection task, including camera view, situation map, generated plan, and execution logs.}
    \label{fig:interface}
\end{figure}

\section{Demonstrations}
\label{sec:demo}
Fig.~\ref{fig:interface} shows the web console used in our demonstration workflows. In a representative area-reconnaissance task, the user issues a natural-language mission such as \textit{``patrol several waypoints, observe each location, and report findings.''} AerialClaw decomposes the mission into executable skills, including takeoff, waypoint navigation, visual observation, status query, and report generation. During execution, the web console displays the generated decisions, execution logs, telemetry, map state, and sensor outputs, allowing users to inspect how the agent selects actions and reacts to feedback.

The demonstrations cover three deployment and reasoning capabilities. First, mock-mode execution reproduces the complete agent loop and web interface without UAV hardware or full simulator dependencies, making it suitable for quick inspection by reviewers and new users. Second, PX4/Gazebo simulation connects the same high-level mission logic to physics-based flight control, sensor-equipped UAV models, and camera/LiDAR streams through the adapter interface. Third, the Reflection Engine summarizes completed task traces into structured lessons, which can be retrieved in later missions to support experience reuse without task-specific reprogramming.

\section{Conclusion and Future Work}
\label{sec:conclusion}
We introduced AerialClaw, an open-source software framework for building LLM-driven autonomous aerial agents. AerialClaw moves UAV applications beyond pre-defined command execution by integrating natural-language mission understanding, closed-loop decision-making, hybrid hard/soft skills, perception feedback, memory, runtime validation, platform adapters, and a web-based control console into a reusable software stack. With mock execution, PX4/Gazebo, AirSim, and pluggable model backends, AerialClaw lowers the engineering barrier for developing UAV systems that can interpret missions, execute skills, and adapt their behavior from feedback. Future work will improve perception grounding, long-horizon memory retrieval, and real-world UAV deployment.

\bibliographystyle{ACM-Reference-Format}
\bibliography{sample-base}

\end{document}